\documentclass{article}

\usepackage{PRIMEarxiv}

\usepackage[utf8]{inputenc} 
\usepackage[T1]{fontenc}    
\usepackage{hyperref}       
\usepackage{url}            
\usepackage{booktabs}       
\usepackage{amsfonts}       
\usepackage{nicefrac}       
\usepackage{microtype}      
\usepackage{lipsum}
\usepackage{fancyhdr}       
\usepackage{graphicx}       
\graphicspath{{media/}}     
\usepackage{cite}
\usepackage{amsmath,amssymb,amsfonts}
\usepackage{algorithmic}
\usepackage{graphicx}
\usepackage{textcomp}
\usepackage{caption}
\usepackage{url}
\title{UnSeGArmaNet: Unsupervised Image Segmentation using Graph Neural Networks with Convolutional ARMA Filters}

\author{
  Kovvuri Sai Gopal Reddy, Bodduluri Saran, A. Mudit Adityaja, Saurabh J. Shigwan, Nitin Kumar,\\ \textbf{and Snehasis Mukherjee}\\
  Shiv Nadar Institution of Eminence Deemed to be University\\
  Delhi-NCR, IN\\
  \texttt{\{kr521,bs404,ma646,saurabh.shigwan,nitin.kumar,snehasis.mukherjee\}@snu.edu.in} \\
}

\begin{document}


\maketitle

\begin{abstract}
The data-hungry approach of supervised classification drives the interest of the researchers toward unsupervised approaches, especially for problems such as medical image segmentation, where labeled data are difficult to get. Motivated by the recent success of Vision transformers (ViT) in various computer vision tasks, we propose an unsupervised segmentation framework with a pre-trained ViT. Moreover, by harnessing the graph structure inherent within the image, the proposed method achieves a notable performance in segmentation, especially in medical images. We further introduce a modularity-based loss function coupled with an Auto-Regressive Moving Average (ARMA) filter to capture the inherent graph topology within the image. Finally, we observe that employing Scaled Exponential Linear Unit (SELU) and SILU (Swish) activation functions within the proposed Graph Neural Network (GNN) architecture enhances the performance of segmentation. The proposed method provides state-of-the-art performance (even comparable to supervised methods) on benchmark image segmentation datasets such as ECSSD, DUTS, and CUB, as well as challenging medical image segmentation datasets such as KVASIR, CVC-ClinicDB, ISIC-2018.
The github repository of the code is available on \url{https://github.com/ksgr5566/UnSeGArmaNet}.
\end{abstract}

\section{Introduction}
\label{sec:intro}
The segmentation of image regions is pivotal in the subject analysis in both healthcare and computer vision domains. A multitude of methods have been proposed to tackle this challenge, harnessing sophisticated techniques primarily in supervised setups. Supervised deep learning-based methods often require a huge amount of annotated data for training, which is often difficult to get, especially in medical image segmentation. We propose an unsupervised method for image segmentation, which is a less explored area of research.

Over the past decade, CNNs have emerged as a popular choice for image segmentation tasks \cite{yolov7}. Prominent architectures rooted in CNNs include UNet \cite{unet}, SegNet \cite{badrinarayanan2017segnet}, and DeepLab \cite{chen2017deeplab}, among others. These models employ multiple layers to extract hierarchical features from the training data. Despite achieving significant success, supervised methods face inherent limitations due to their reliance on labeled datasets. As building extensive annotated datasets is resource-intensive and often infeasible due to privacy constraints, especially in healthcare and surveillance. Additionally, supervised models trained on specific datasets may struggle to generalize to new or diverse datasets due to variations in imaging protocols, equipment, and subject types. This lack of generalization can constrain the broader applicability of the segmentation model. Recently, MedSAM \cite{ma2024segment}, known as the Segment-Anything Model in the medical imaging domain, represents a step towards achieving generalization in medical image segmentation. This model employs a vision transformer (ViT)-based architecture for both image encoding and decoding. By combining a pre-trained Segment-Anything (SAM) \cite{sam} model with the ViT-base \cite{caron2021emerging} model, MedSAM undergoes fully supervised training on an extensive dataset.

However, MedSAM faces a challenge related to modality imbalance within the dataset \cite{ma2024segment}, where certain structures or pathologies may have been underrepresented. As a result, the segmentation model demonstrates a bias towards the majority class, leading to suboptimal performance for minority classes. Moreover, it has been only trained and experimented with cancer data. Further, these ViT-based techniques often ignore the inherent graph structure of the objects in the image. Recently, the combination of ViT and Graph Convolution Network (GCN) has been applied to computer vision datasets, providing promising results \cite{tanishka}. Efforts have been made to apply ARMA filters with GCN, for feature aggregation \cite{bianchi2021graph}.

In this work, we apply a ViT network for extracting features from the images. We first form a complete graph structure from the extracted features from which, we prune out some edges based upon a similarity measure between objects. We apply the ARMA filter for feature aggregation. Finally, we apply a Modularity loss for unsupervised optimization of segmentation clusters following \cite{tsitsulin2023graph}. The contributions of this paper can be summarized as follows:
\begin{itemize}
\item We propose a cross-field unsupervised segmentation framework that harnesses the inherent graph structure within input images, with features extracted from a pre-trained ViT. This is an attempt to leverage the benefits of features extracted by a ViT, and the inherent graph structure of the image, and apply it for image segmentation.
\item We introduce a modularity-based loss function for image segmentation, coupled with an auto-regressive moving average (ARMA) filter \cite{isufi2016autoregressive,chen_icml20}. This combination captures the underlying graph topology within the images.
\item We observe that employing SILU and SELU activation functions within the proposed Graph Neural Network results in enhancing the segmentation accuracy. 
\end{itemize}
\begin{figure*}
    \centering    \includegraphics[width=0.9\linewidth,trim={0 1cm 0 0}]{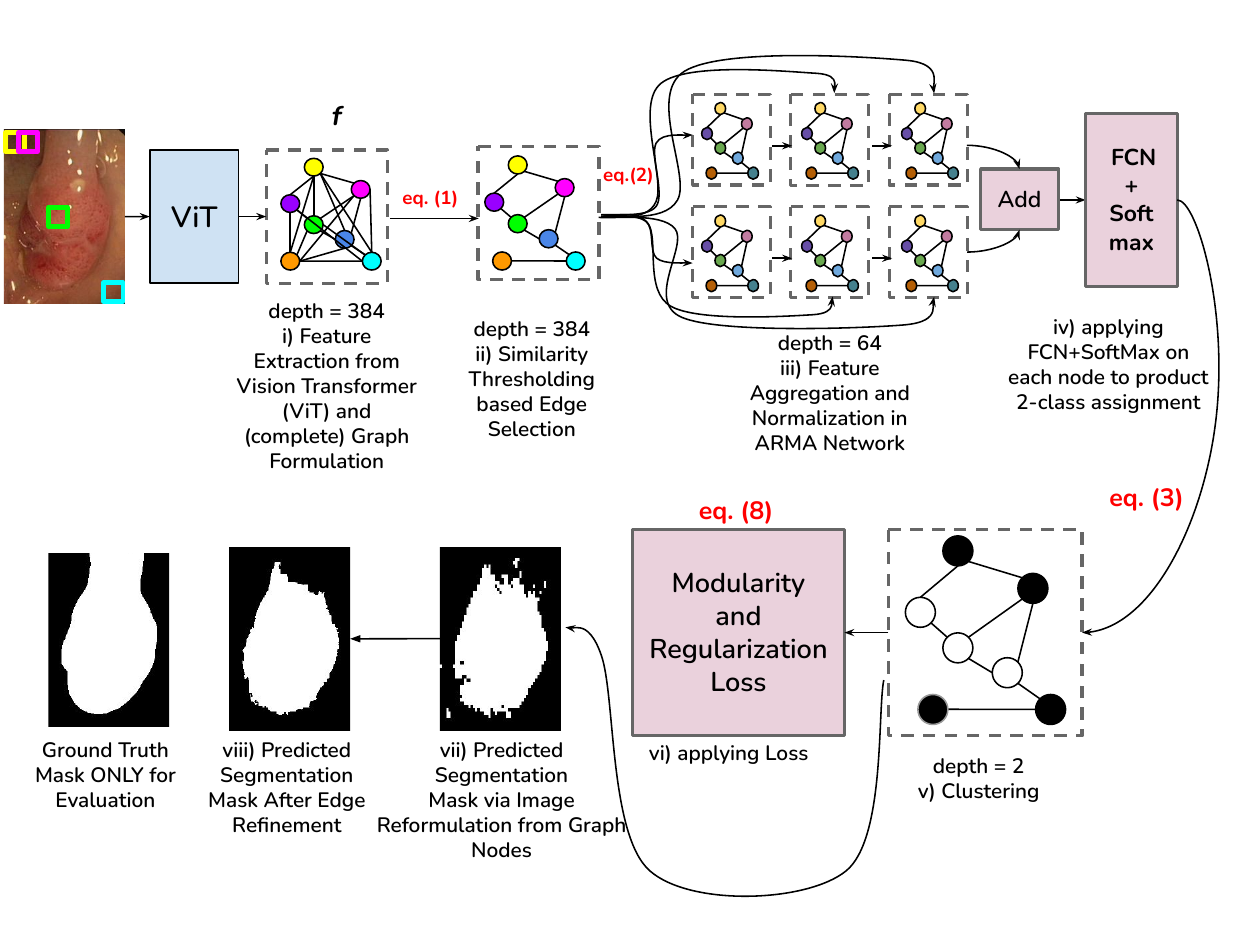}
    \caption{UnSegArmaNet Pipeline: we i) extract features $f$ of all (overlapping) image patches using vision transformer (ViT) and formulate a (complete) Graph $G$ (few nodes shown, for illustration, in the same color as image patch windows), ii) then apply similarity (normalized $ff^T$) threshold to select important edges in $G$, iii) aggregate and normalize features in graph Arma network (ArmaConv), darker node colors represent aggregation, iv) apply a fully connected network (FCN) to finally obtain node level clusters. vi) The modularity and regularization-based loss is finally used to train the model. vii-viii) At inference, edge refinement is used over the predicted mask.}
    \label{fig:method}
\end{figure*}

\section{Related Works}
Image segmentation, especially for medical images, remains a challenge despite several efforts ranging from traditional methods to deep CNNs and ViTs \cite{survey}. During the last decade, several CNN models have shown comparative performances in segmenting medical images, as well as general images \cite{badrinarayanan2017segnet,unet,chen2017deeplab}. Segnet \cite{badrinarayanan2017segnet} is an encoder-decoder network, followed by a pixel-wise semantic classification layer for segmentation. In order to handle the varying scales of the objects in images, deeplab \cite{chen2017deeplab} utilizes atrous convolution operation for semantic segmentation. The CNN based methods require a huge amount of annotated data for training, which makes them difficult to be applied to medical images. To address this issue, UNet architectures is widely used for medical image segmentation due to its ability to augment data \cite{unet}. However, extracting the minute texture information remains a challenge.

Apart from vision-related tasks, ViT models gained popularity in biomedical image segmentation due to the ability to extract minute information from images, with the inherent attention mechanism \cite{survey_vit}. MinCutPool \cite{bianchi2020spectral} presents a pooling policy based on the relaxation of the normalized min-cut problem \cite{shi2000normalized}. This approach utilizes spatially localized graph convolutions in graph neural networks, avoiding the need for expensive spectral decomposition. In contrast, DSM \cite{melas2022deep} encodes similarities using the correlation of ViT features and color information in the graph adjacency matrix. However, this approach requires performing spectral decomposition on the adjacency matrix to segment the input image, which is computationally expensive. While DINO-ViT \cite{caron2021emerging,paul2022vision} generates useful global features, it often leads to coarse image segmentation.

Efforts have been made to combine the ViT features with GCN, to extract the inherent graph structure of the image \cite{aflalo2023deepcut}. However, the data-hungriness of supervised algorithms remains a challenge. Fewer efforts are made for unsupervised image segmentation, especially in the biomedical domain. GDISM \cite{trombini2023goal} is an unsupervised image segmentation approach that combines parametric and graph-based techniques to generate image segments aligned with user-defined application-specific goals. It initializes a set of points called seeds and then identifies homogeneous and non-granular regions by minimizing a custom energy function combining the graph-based methods and Markov random fields. While GDISM effectively captures local contextual information, its underlying methodologies result in limited mean intersection over union (mIoU) performance.

Our proposed UnSeGArmaNet framework is an unsupervised model that effectively leverages the features to capture both local and global contexts in images from diverse modalities. Additionally, UnSeGArmaNet is a shallow network that requires minimal time and effort to segment input images.

\section{Proposed Method}
Fig. \ref{fig:method} shows the overall procedure of the proposed method. We use features from a ViT \cite{caron2021emerging} that breaks down images into fixed-size patches, making it more efficient and scalable than traditional CNNs by capturing global context in images using a self-attention mechanism \cite{ramachandran2019stand,zhao2020exploring}. The ViTs are less sensitive to data augmentation compared to CNNs \cite{paul2022vision}.
\\
Our method takes an input image of size $s\times t$ with $d$ channels and passes it through a pre-trained small vision transformer (ViT) without fine-tuning. The transformer divides the image into patches of size $p \times p$, resulting in $st/p^2$ patches. We extract the internal representation of each patch from the key layer features of the final transformer block, which has shown excellent performance across various tasks. This produces feature vectors of size $(st/p^2) \times C_{in}$, where $C_{in}$ is the token embedding dimension for each image patch.
This is followed by the creation of an input image-specific graph $G$ which captures the neighborhood relationships using correlation $A$ defined by
\begin{equation}
\label{eq:thresh_corr}
A= \biggr(ff^T > \tau\biggr) \in \mathbb{R}^{\frac{st}{p^2}\times \frac{st}{p^2}},
\end{equation}
where individual entries of adjacency matrix $A$, $A_{ij} \in \{0,1\}$ and $\tau  \in (0,1)$ is a user-defined parameter which is tuned specific to the dataset. The illustrative result of applying equation (\ref{eq:thresh_corr}) on the normalized node features $f$ of $G$ is shown in Fig.~\ref{fig:method} (ii).

Here, the graph nodes represent the image patches. Different patch windows and their corresponding nodes are shown in common colors in Fig.~\ref{fig:method}'s input image and Fig.~\ref{fig:method} (i).
\\
We employ a Graph Neural Network (GNN) \cite{scarselli2008graph}, which relies on graph convolutional layers to process the data. These layers enable localized spectral filtering~\cite{kipf2016semi}, allowing us to capture the relationships between nodes that represent image patches. We measure these neighborhood relationships by calculating the thresholded correlation between the feature vectors $f$ of adjacent nodes using (\ref{eq:thresh_corr}).
\\
The graph convolution skip layers\cite{lu2024skipnode} of  UnSeGArmaNet at $r^{th}$ row stack are:
\begin{equation}
\bar{X}_r^{(l+1)} = \sigma(\hat{A}\bar{X}_r^{l}W_{(r)}^{(l)} + XV_{(r)}^{(l)}),
\label{eq:Arma_Conv}
\end{equation}
where $W_{(r)}^{(l)}, V_{(r)}^{(l)}$ are the learnable parameters at $l^{th}$ layer at $r^{th}$ row stack, $X$ is the initial node feature. Here $\hat{A} = D^{\frac{-1}{2}}AD^{\frac{-1}{2}}$, where $D \in \mathbb{R}^{\frac{st}{p^2}\times \frac{st}{p^2}}$ is a diagonal degree matrix with $D_{ii} = d_i$, in which $d_i$ represents degree of $i^{th}$ node. In ArmaConv layer\cite{bianchi2021graph}, there are $R$ row stacks each containing $L$ layers. Output of the ArmaConv layer $\bar{\bar{X}} = \frac{1}{R}\sum_{r=1}^{R}\bar{X}_r^{(L)}$, is passed to a FCN layer and softmax to get final cluster assignment $C \in[0, 1] ^{n\times k}$. 
Here, $n,k$ are the number of nodes and clusters respectively as mentioned in equation (\ref{eq:cluster_C}) (see Fig. \ref{fig:method} (iv), where a black node represents 0 class, and a white node represents 1 class).
\begin{equation}
 C = softmax(MLP(ArmaConv(\hat  {A},X))).
 \label{eq:cluster_C}
\end{equation}

The loss function relies on the modularity matrix $B$ which is defined on an undirected graph $\mathcal{G}= (\mathcal{V},\mathcal{E})$ where $\mathcal{V}=(v_1,\ldots,v_n),$ is the set of $n$ nodes and edges $\mathcal{E} \subseteq \mathcal{V} \times \mathcal{V}$. Let $A$ be the adjacency matrix of the undirected graph $\mathcal{G}$ such that $A_{ij}=1 \hspace{3pt}$ when $\hspace{3pt}\{v_i,v_j\} \in \mathcal{E}$ and $A_{ij}=0$ otherwise, then: 
\begin{equation}
\label{eq:mod_matrix}
B= A - \frac{dd^T}{2m}.
\end{equation}

The modularity matrix $B$ is defined in terms of the degree vector $d$ of the graph $\mathcal{G}$ and the total number of edges  $m=|\mathcal{E}|$. The matrix $B$ captures the differences between the observed edges in $\mathcal{G}$ and the expected edges in a random graph with the same degree sequence. Positive values in $B$ indicate a higher density of edges within a cluster, contributing to a higher modularity score. The modularity $Q$ of a partition $C$ of the graph can be calculated as:
\begin{equation} 
\label{eq:modularity}
Q = \frac{1}{2m}\sum_{ij}\bigg[A_{ij}-\frac{d_i d_j}{2m}\bigg]\delta(c_i,c_j),
\end{equation}
where $c_i$ represents the cluster assignment of node $i$, and $\delta$ is the Kronecker delta function:
\begin{equation}
\delta(a, b) = \begin{cases} 1, & \text{if } a = b \\ 0, & \text{if } a \neq b \end{cases}.
\end{equation}
A higher modularity score indicates a better partitioning of the graph $\mathcal{G}$, so maximizing the modularity $Q$ is equivalent to achieving better clustering. However, it has been proven that maximizing equation (\ref{eq:modularity}) is an NP-hard problem~\cite{tsitsulin2023graph}. Therefore, a relaxed version of the equation, (\ref{eq:rel_bar_Q}), is typically used for optimization purposes.
\begin{equation}
\label{eq:rel_bar_Q}
\Bar{Q}= \frac{1}{2m}Tr(C^TBC),
\end{equation}
where the cluster assignment matrix $C \in\mathbb{R} ^{n\times k}$ ($n,k$ are the number of nodes and clusters respectively) is computed using equation (\ref{eq:cluster_C}). The trace operator, denoted by $Tr(\cdot)$, calculates the sum of the diagonal elements of a square matrix. We cluster the resulting patch features from ViT by minimizing the loss function $\mathcal{L}$, as defined in equation (\ref{eq:loss_fun}). This optimization process ensures that the resulting cluster assignment matrix C is non-trivial, as proven in~\cite{tsitsulin2023graph}).
\begin{equation}
 \label{eq:loss_fun}
 \mathcal{L} = -\frac{1}{2m}Tr(C^TBC) + \frac{\sqrt{k}}{n}\bigg\lVert \sum_{i=1}^{n} C_i \bigg\rVert_F - 1.
 \end{equation}
 The variable $C_i$ represents the soft cluster assignments for the $i^{th}$ node, where each assignment is a vector of values between $0$ to $1$ with the length of $k$. The notation $\lVert.\rVert_F$ denotes the Frobenius norm, a mathematical operation used to measure the magnitude of a matrix.
 
\section{Experiments, Results and Discussions}
We first detail our experimental setup, followed by results and discussions.

\subsection{Experimental Setup}
We start with a brief description about the different datasets used in our study, followed by the implementation details.
\subsubsection{Datasets}
\textbf{CVC-ClinicDB~\cite{bernal2015wm}}
is a well-known open-access dataset for colonoscopy research. It was introduced as part of the Computer Vision Center (CVC) Colon Image Database project. It contains 612 images with a resolution of 384×288 from 31 colonoscopy sequences acquired during routine colonoscopies.\\
\textbf{KVASIR~\cite{jha2020kvasir}} includes images from the gastrointestinal tract, obtained through various endoscopic procedures such as gastroscopy and colonoscopy. The dataset contains 8,000 endoscopic images, with 1,000 image examples per class.  Images in the KVASIR dataset are typically labeled with 8 different classes related to gastrointestinal conditions, lesions, or abnormalities. \\
\textbf{ISIC-2018~\cite{codella2019skin}}
The ISIC (International Skin Imaging Collaboration) 2018 is a dataset and challenge focused on dermatology and skin cancer detection. The dataset is designed to facilitate research in the development of computer algorithms for the automatic diagnosis of skin lesions, including the identification of melanoma. The dataset includes clinical images, dermoscopic images, and images obtained with confocal microscopy.\\
\textbf{ECSSD~\cite{shi2015hierarchical}}  ECSSD (Extended Complex Scene Stereo Dataset) is a dataset commonly used  to identify the most visually significant regions within an image. It has been designed to evaluate the performance of algorithms on complex and natural scenes. The dataset contains 1000 RGB images captured in complex and diverse scenes along with the annotated pixel-wise ground-truth masks indicating the saliency regions created as an average of the labeling of five human participants.\\
\textbf{DUTS~\cite{wang2017learning}} 
The DUTS (Deep Unsupervised Training for Salient Object Detection) dataset is widely used in the field of computer vision for the evaluation of salient object detection algorithms. It contains 10,553 training images and 5,019 test images. 
The dataset consists of images with diverse and complex scenes and each image is annotated with a pixel-wise ground truth mask which indicates the salient object within the scene. We use only the test dataset for evaluation.\\
\textbf{CUB~\cite{wah2011caltech}} CUB (Caltech-UCSD Birds) dataset is a widely used dataset in the field of computer vision, particularly for the classification of bird species. It contains images of 200 bird species, with a total of around 11,788 images. Each image is associated with a specific bird species. The dataset provides detailed annotations
for bird parts such as head, body, and tail.
\subsubsection{Implementation Details}
We employ the DINO-ViT small model \cite{caron2021emerging} with a patch size of $8$, resulting in node features of size $C_{in}=384$. We then utilize a two-stacks, four-layered Graph ARMA Convolutional Network (Graph ARMAConv \footnote{ARMAConv is a module defined in torch\_geometric library}) to aggregate these features, followed by two fully connected network (FCN) projections: first from $384$ to $64$, and then from $64$ to $k$ (where $k = 2$) for binary segmentation. The \textit{UnSeGArmaNet} model is individually optimized for each image using the loss function $\mathcal{L}$. The initial learning rate is configured to $10^{-3}$ with a decay factor of $10^{-2}$. We employ the ADAM~\cite{kingma2014adam} for the optimization of $\mathcal{L}$ in $60$ epochs. Through experimentation, we find that setting the threshold parameter $\tau$ within the range $0.4 \leq \tau \leq 0.6$ yields improved accuracy and faster convergence across various datasets. Additionally, we explore different activation functions such as SiLU~\cite{elfwing2018sigmoid}, SeLU~\cite{ramachandran2017searching}, GELU~\cite{hendrycks2016gaussian} and ReLU~\cite{glorot2011deep}, observing that SiLU and SeLU are particularly effective for our application.

\subsection{Results}
\label{sec:exp_det}
When comparing our UnSeGArmaNet model to medical image datasets, we used the following methods:  MedSAM \cite{ma2024segment}, MinCutPool\cite{bianchi2020spectral}, GDISM\cite{trombini2023goal}, DSM\cite{melas2022deep} and ViT-Kmeans. As shown in Table I, UnSeGArmaNet outperforms state-of-the-art (SOTA) unsupervised methods. Notably, UnSeGArmaNet achieves significantly better scores on ISIC-2018 compared to MedSAM. However, MedSAM performs well on KVASIR and CVC-ClinicDB due to its training on large polyp datasets. On the other hand, MedSAM struggles with ISIC-2018 due to modality-imbalance issues. We also explored the impact of different activation functions on UnSeGArmaNet, including SiLU, SeLU, GELU, and ReLU. The results are presented in Table III. UnSeGArmaNet with SiLU activation function yields the best results on all the four medical image datasets, while SeLU activation function provides the best results on other benchmarks, surpassing SOTA unsupervised methods.
We evaluate the segmentation results using the Mean Intersection over Union (mIOU) score, which is based on binarized values. The mIOU score is calculated as the average of the IoU scores across all the $N$ classes, and is defined as:
\begin{equation}
\label{eq:miou}
    mIOU = \frac{1}{N}\sum_{i=1}^{N}IoU_i,
\end{equation}
where IoU for class $i$ is:
\begin{equation}
\label{eq:iou}
    IOU_i = \frac{TP_i}{TP_i + FP_i + FN_i},
\end{equation}
where $TP_i$ are the true positives i.e. the number of pixels correctly classified as class $i$; $FP_i$ are the false positives, or the number of pixels incorrectly classified as class $i$ and $FN_i$ are the false negatives, or the number of pixels incorrectly classified as a different class other than class $i$. The mIOU score provides a comprehensive evaluation of a model's segmentation performance by assessing the overlap between predicted and ground truth masks for each class. A higher mIOU value indicates better segmentation accuracy, with higher values indicating better performance.
\begin{table*} [t]
\centering
\setlength{\tabcolsep}{2pt}
\caption{Average mIOU scores for medical image data}
\resizebox{\textwidth}{!}{%
\begin{tabular}{l|ccccccc}
 \textbf{Datasets}
& \textbf{Ours+ARMA} & \textbf{Ours+GCN}& \textbf{MedSAM}~\cite{ma2024segment} & \textbf{MinCutPool~\cite{bianchi2020spectral}} & \textbf{GDISM\cite{trombini2023goal}} & \textbf{DSM}~\cite{melas2022deep} & \textbf{ViT-Kmeans} \\ 
\midrule \hline
KVASIR & 75.23 & 74 & 76.72 & 73.50 & 59.40 & 58.80 & 66.10 \\
ISIC-2018 & 68.72 & 73.94 & 61.36 & 72.31 & 52.53 & 72.20 & 68.60  \\
CVC-ClinicDB & 67.93 & 64 & 71.53 & 62.10 & 59.22 & 60.48 & 62.80  \\
\end{tabular}
}
\label{table:tab1}
\end{table*}
\begin{table}[t]
\centering
\setlength{\tabcolsep}{2pt}
\caption{Average mIOU scores for computer vision datasets}
\resizebox{0.8\textwidth}{!}{%
\begin{tabular}{l|ccccc}
\textbf{Datasets}
& \textbf{Ours+ARMA} & \textbf{Ours+GCN}  & \textbf{MinCutPool\cite{bianchi2020spectral}} & \textbf{DSM}\cite{melas2022deep} & \textbf{ViT-Kmeans} \\ 
\midrule \hline
ECSSD & 77.83 & 75.57 & 74.6 & 73.3 & 65.76\\
DUTS & 60.3& 61.04 & 59.5 & 51.4 & 42.78\\
CUB & 81.23 & 78.41 & 78.2 & 76.9 & 51.86\\
\end{tabular}
}
\label{table:tab2}
\end{table}

\begin{figure*}[ht!]
    \centering
    \includegraphics[width=\linewidth,trim={0 0cm 0cm 0},clip]{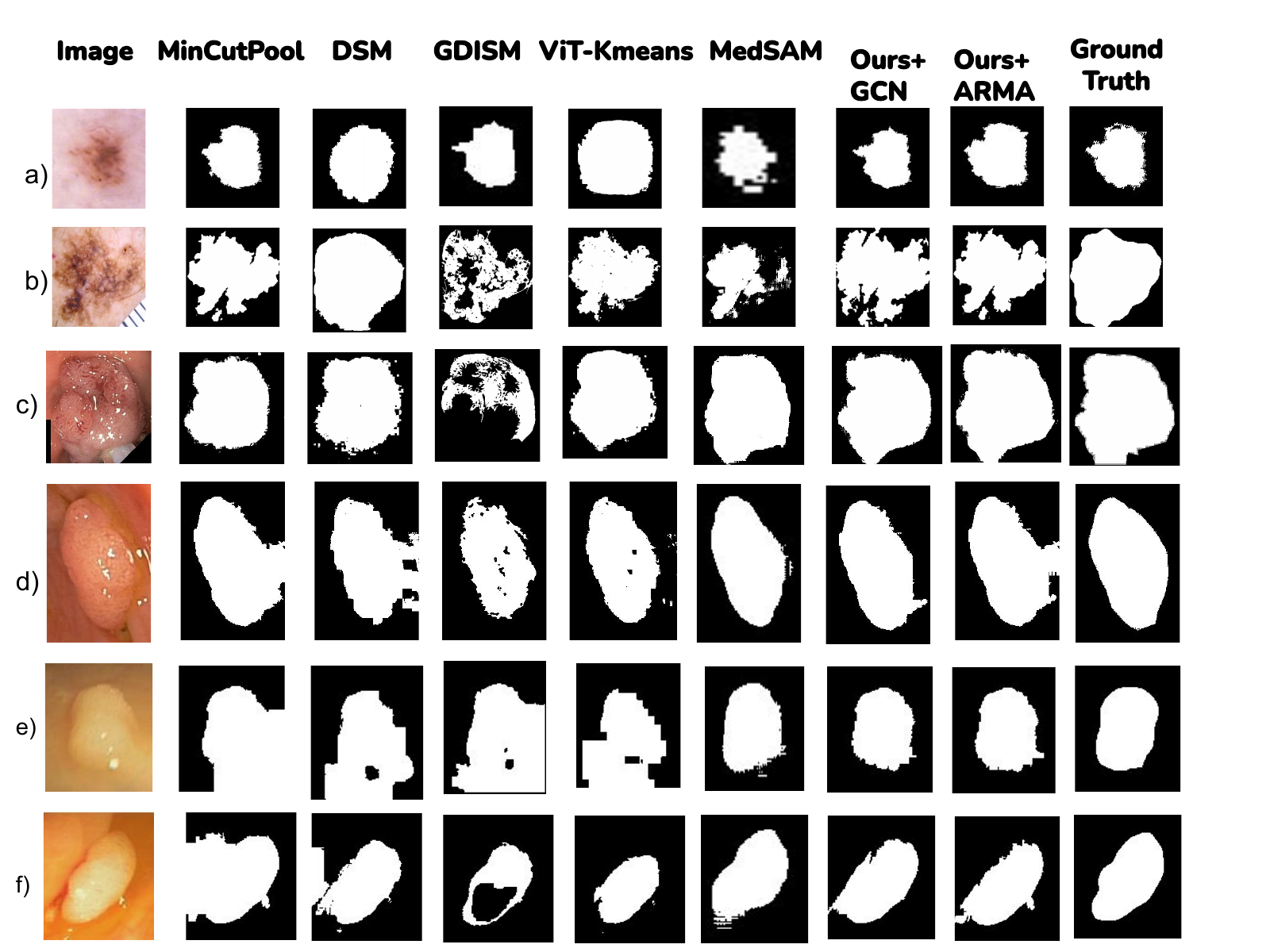}
    \caption{Segmentation results on  (a)-(b) ISIC-2018,  (c)-(d) KVASIR, (e)-(f) CVC-ClinicDB sample images}
    \label{fig:enter-label}
\end{figure*}
\begin{figure*}[ht!]
    \centering
    \includegraphics[width=\linewidth,trim={0 0cm 0cm 0},clip]{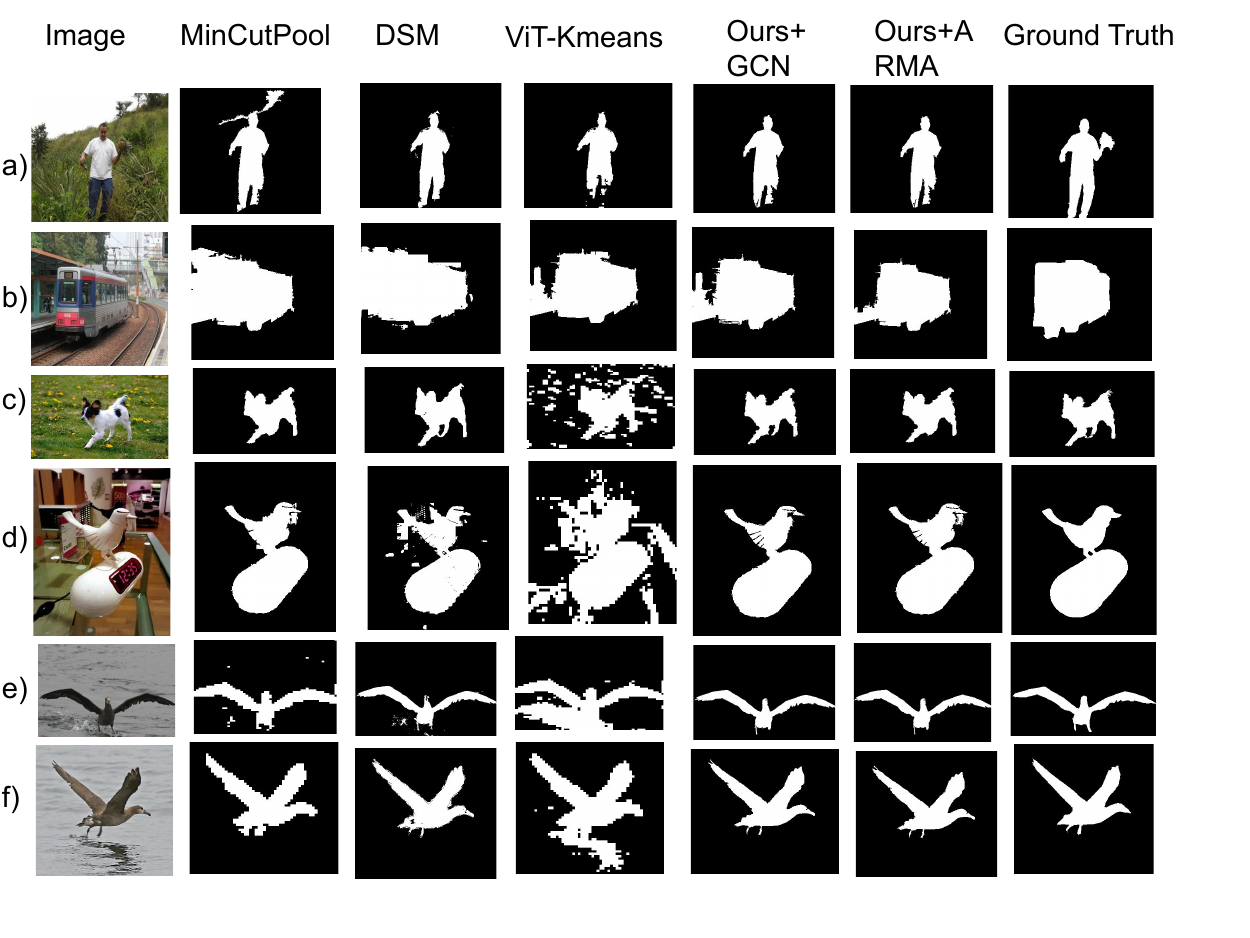}
    \caption{Segmentation results on  (a)-(b) ECSSD,  (c)-(d) DUTS, (e)-(f) CUB sample images}
    \label{fig:cv_results}
\end{figure*}
\subsubsection{Results on Medical Image Datasets}
As shown in Table~\ref{table:tab1}, UnSeGArmaNet outperforms other unsupervised methods. Notably, it achieves comparable results to MedSAM on KVASIR~\cite{jha2020kvasir} and ISIC-2018~\cite{codella2019skin} datasets, despite MedSAM being fine-tuned on a large pool of medical image datasets, including large colonoscopy datasets. However, MedSAM struggles with modality-imbalance issues, which may have contributed to its better performance on KVASIR and CVC-ClinicDB datasets, but significantly lower scores on ISIC-2018. This imbalance is likely due to MedSAM's training on large polyp and small dermoscopy image-mask pairs.

\subsubsection{Results on Other Vision Datasets}
We evaluate UnSeGArmaNet on publicly available ECSSD, CUB, and DUTs vision datasets, comparing it to MinCutPool~\cite{bianchi2020spectral}, DSM~\cite{melas2022deep}, and ViT-Kmeans. The results, summarized in Table \ref{table:tab2}, show that UnSeGArmaNet outperforms the other methods. Notably, UnSeGArmaNet and the compared methods all utilize ViT to extract high-level features. However, ViT-Kmeans, which applies k-means algorithm~\cite{kanungo2002efficient} to these features, achieves significantly lower scores than UnSeGArmaNet on all datasets. This highlights that high-level features from VIT alone are insufficient for effective segmentation. Our modularity-based UnSeGArmaNet framework surpasses the performance of ~\cite{bianchi2020spectral} and ~\cite{melas2022deep}, which employ graph-cut based spectral decomposition in their deep learning architectures.
\subsection{Discussions}
Our approach consists of two components: a pre-trained network and a Graph Arma Convolutional Neural Network (ARMAConv) with its optimization. We utilize DINO-ViT small as the pre-trained network, which is a widely adopted feature extractor known for providing robust and adaptable features across different data modalities \cite{paul2022vision}. However, as demonstrated through the results of ViT-Kmeans, (where ViT features are fed to vanilla k-means algorithm), these features alone are insufficient for producing high-quality segmentation results. This is evident from the coarse segmentation outcomes shown in Fig. \ref{fig:enter-label} and Fig. \ref{fig:cv_results}. The idea of GCN integrated with ARMA convolutional filter is applied in MedSAM \cite{ma2024segment}. Although MedSAM is a generalized model for universal medical image segmentation, it is vulnerable to modality imbalance issues, as discussed in Section~\ref{sec:exp_det}. Moreover, even if images from all the modalities are equally represented, it does not guarantee good accuracies across all the modalities. Furthermore, updating such a model is computationally expensive, making it a challenging task. While DINO-ViT generates useful global features, it often leads to coarse image segmentation, as depicted in Fig. \ref{fig:enter-label} and Fig. \ref{fig:cv_results}. Whereas, the proposed model can effectively leverage the features to capture both local and global contexts in images from diverse modalities. While GDISM \cite{trombini2023goal} effectively captures local contextual information (Fig.\ref{fig:enter-label} and Fig. \ref{fig:cv_results}), its underlying methodologies result in limited mean intersection over union (mIoU) performance as shown in Table \ref{table:tab1}.

\section{Conclusions}
We proposed an unsupervised image segmentation procedure capable of combining the benefits of the attention-based features captured by a ViT, and the inherent graph structure of the image captured by a GCN model. Further, we applied ARMA filter for feature aggregation. Our experiments on the benchmark datasets show state-of-the-art results, both on medical image datasets and vision datasets for image segmentation. The proposed model is lightweight and hence, is less prone to overfitting problems. Further, the proposed unsupervised model provides a useful way to solve various tasks in biomedical image analysis, where getting annotated data is difficult. In the future, the proposed GCN-based model over ViT features can be tested on other challenging medical images such as histopathology images, or even for medical image classification, object recognition in medical images, and several other medical imaging problems. 

\section*{Declarations}
\textbf{Conflict of interest}
The authors declare that they have no conflict of interest.\\
\textbf{Data availability}
All datasets used for our experiments are freely available.
\bibliographystyle{unsrt}  
\bibliography{egbib}

\end{document}